\documentclass[conference]{IEEEtran}
\IEEEoverridecommandlockouts
\usepackage{cite}
\usepackage{amsmath,amssymb,amsfonts}
\usepackage{graphicx}
\usepackage{textcomp}
\usepackage{xcolor}
\usepackage{gensymb }

\usepackage{graphicx}
\usepackage{subcaption}
\usepackage{multirow}
\usepackage{booktabs}       
\usepackage{dsfont}
\usepackage{hyperref}

\usepackage{algorithm}
\usepackage{algorithmicx}
\usepackage[noend]{algpseudocode}
\algnewcommand{\LineComment}[1]{\Statex \(\triangleright\) #1}
\algnewcommand\algorithmicinput{\textbf{Input:}}
\algnewcommand\algorithmicoutput{\textbf{Output:}}
\algnewcommand\Input{\item[\algorithmicinput]}
\algnewcommand\Output{\item[\algorithmicoutput]}

\def\algbackskip{\hskip-\ALG@thistlm}

\algdef{SE}[FUNCTION]{Function}{EndFunction}%
   [2]{\algorithmicfunction\ \textproc{#1}\ifthenelse{\equal{#2}{}}{}{(#2)}}%
   {\algorithmicend\ \algorithmicfunction}%

\def\BibTeX{{\rm B\kern-.05em{\sc i\kern-.025em b}\kern-.08em
    T\kern-.1667em\lower.7ex\hbox{E}\kern-.125emX}}

\begin{document}

\title{Correlation and Unintended Biases on Univariate and
Multivariate Decision Trees
}

\author{\IEEEauthorblockN{Mattia Setzu}
\IEEEauthorblockA{\textit{Department of Computer Science} \\
\textit{University of Pisa}\\
Pisa, Italy \\
\href{mailto:mattia.setzu@unipi.it}{mattia.setzu@unipi.it}}
\and
\IEEEauthorblockN{Salvatore Ruggieri}
\IEEEauthorblockA{\textit{Department of Computer Science} \\
\textit{University of Pisa}\\
Pisa, Italy \\
\href{mailto:salvatore.ruggieri@unipi.it}{salvatore.ruggieri@unipi.it}}
}

\maketitle

\begin{abstract}
Decision Trees are accessible, interpretable, and well-performing classification models. 
A plethora of variants with increasing expressiveness has been proposed in the last forty years.
We contrast the two families of \textit{univariate} DTs, whose split functions partition data through axis-parallel hyperplanes, and \textit{multivariate} DTs, whose splits instead partition data through oblique hyperplanes.
The latter include the former, hence multivariate DTs are in principle more powerful.
Surprisingly enough, however, univariate DTs consistently show comparable performances in the literature.
We analyze the reasons behind this, both with synthetic and  real-world benchmark datasets. Our research questions test whether the pre-processing phase of removing correlation among features in datasets has an impact on the relative performances of univariate vs multivariate DTs. We find that existing benchmark datasets are likely biased towards favoring univariate DTs.
\end{abstract}


\section{Introduction}

Decision Trees (DTs)~\cite{DBLP:books/sp/datamining2005/RokachM05} are accessible, interpretable, and well-performing classification models that are commonly used in practice. 
These models are widely available across software libraries and are standard baselines in industry and academic communities \cite{Rudin2016_KeyNotePapis}. DTs are inherently transparent \cite{Rudin2019_StopExplainingML}, which may facilitate the inclusion of stakeholders in understanding and assessing model behaviour. Moreover, ensembles of DTs outperform deep learning models on tabular data~\cite{DBLP:conf/nips/GrinsztajnOV22}.
Trending research topics include the design of DT learning algorithms able to reproduce the behavior of complex and opaque models (e.g., of a tree ensemble \cite{DBLP:journals/jbd/WeinbergL19,DBLP:conf/icml/VidalS20,DBLP:conf/dis/BonsignoriGM21,DBLP:conf/ijcai/DudyrevK21}), and the design of DT learning algorithms able to adapt to distributions of data different from the one of the training dataset \cite{DBLP:conf/fat/0002SBR23}.

The condition tested at internal nodes of a decision tree can be broadly categorized as \emph{univariate} or \emph{multivariate}, the former yielding axis-parallel splits, and thus producing univariate DTs (UDTs), and the latter yielding oblique splits, and thus producing multivariate DTs (MDTs).
These two families of DTs have found wildly different degrees of success in practical use, and in the scientific research.
Multivariate DTs are, in principle, more expressive than the univariate ones, being capable of linearly separating the space of instances. Such an expressivity comes at a higher computational cost and an interpretational gap. Surprisingly enough, when compared head-to-head~\cite{DBLP:journals/ijprai/YildizA05}, there is not a clear-cut performance difference between the two types of DTs. For this, MDTs are often dismissed in favor of univariate ones -- an application of the Occam's razor.

    
In this paper, we analyze the reasons behind this unexpected behavior by analytically comparing the two families of DTs, and by identifying a bias in standard dataset pre-processing practices that favors UDTs.
We further support our theoretical results with an extensive experimentation on both synthetic and real-world datasets, which confirms the above bias.

The paper is structured as follows. In Section~\ref{sec:background}, we define background notions on DTs.
In Section~\ref{sec:method}, we pose our research questions. Experimental results are discussed in Section~\ref{sec:exp}. Finally, in Section~\ref{sec:conclusions}, we summarize our conclusions.

  \section{Background}
    \label{sec:background}

        \subsection{Decision Trees}
        A decision tree $T$ is a tree data structure comprised of $N$ nodes $\{n_i\}_{i = 1}^N$, either internal nodes or leaves.
        Each internal node points to a number of child nodes, the degree of the node, based on the possible outcomes of the split function at the node. Leaf nodes terminate the tree and are associated with task-dependent values, e.g., in a classification task, a leaf holds an estimated probability distribution over class labels.
        Top-down tree learning algorithms, such as 
        CART~\cite{DBLP:books/wa/BreimanFOS84} and C4.5~\cite{DBLP:journals/ml/Quinlan86}, 
        follow a greedy approach, 
        in which the tree structure and the split functions are learned recursively. Optimal decision tree learning algorithms~\cite{DBLP:journals/ml/BertsimasD17} compile a complete binary tree of given depth to a Mixed Integer Linear Programming (MILP) problem. For uniformity, in the former approach, we also restrict  to a tree depth stopping criteria and to binary split functions. Moreover, we disregard pre-processing steps, such as feature selection~\cite{DBLP:journals/jmlr/Ruggieri19}, and post-processing steps, such as tree simplification~\cite{DBLP:journals/pami/EspositoMS97}. For a recent survey on DTs, see \cite{DBLP:journals/air/CostaP23}.


    

    \subsection{Split functions}
    We assume binary split functions (or, simply, splits) $f_i: \mathcal{X} \rightarrow \{0, 1\}$, that at an internal node $n_i$ deterministically redirect an instance $x \in \mathcal{X} \subseteq \mathrm{R}^m$ towards the left ($f_i(x)=0$) or the right ($f_i(x)=1$) child node. Some approaches learn a fuzzy split condition~\cite{adamo1980fuzzy}, for which the instance is probabilistically redirected to one of the child nodes. 
%
%
    Split functions can vary in complexity, from being as simple as a univariate split~\cite{DBLP:books/wa/BreimanFOS84,DBLP:journals/ml/Quinlan86} or as complex as a neural network~\cite{DBLP:conf/ijcai/KontschiederFCB16}.
    The most common split functions are of the form \textit{1-of-1} splits, that is, they are comprised of a single boolean condition, for which instances are redirected according to the result of the condition. 
    In multiconditional DTs, such as Trepan~\cite{DBLP:conf/nips/CravenS95}, split functions consist of testing $k$ boolean conditions, $h$ of which must be satisfied -- an $h$-of-$k$ split policy.
    This introduces a significant cost in terms of both learning complexity and interpretability. 
%
    In the rest of this paper, we restrict to uniconditional \emph{linear} split functions of the form:
    \begin{equation}\label{eq:splits}
    f_i(x) = \mathds{1}(\alpha_i^T x \leq \beta_i)
    \end{equation}
    where $\alpha_i \in \mathds{R}^m, \beta_i \in \mathds{R}$ are the parameters of the split function $f_i$.
    Univariate splits implement univariate separating hyperplanes, that is, $\alpha_i$ is $0$ but for a single component for which it is $1$.
    Multivariate splits allow for more than one non-zero component, hence implementing separating hyperplanes with an arbitrary number of non-zero components.


    \subsection{Univariate splits}
      Let us denote by $n_0$, $n_1$, and $n_2$ a parent node and its two child nodes respectively. 
      Also, let $X_i$ and $Y_i$, for $i=0,1,2$, be the features and class attributes respectively, of the data instances at the three nodes.
      Univariate splits test a condition $x_j \leq \beta$, where $x_j$ with $j \in 1, \ldots, m$ is a feature in $X_0$ and 
      and $\beta$ is in the domain of $x_j$.
      According to~\cite{DBLP:series/sci/2014-498}, we distinguish the following strategies.

      \textit{Impurity-based splits.}
        Impurity measures the  heterogeneity of class label frequency for instances at a node.
        We can define vectors $p_1$ and $p_2$ holding the empirical class distribution of the instances at $n_1$ and $n_2$, i.e., called $\mathit{freq}(y, Y_i)$ the frequency of class label $y$ in $Y_i$:
        \begin{equation}
          p_1 = \left\langle \dfrac{\mathit{freq}(y, Y_1)}{\mid Y_1 \mid}  \right\rangle_{y \in \mathcal{Y}} \hspace{0.5cm} p_2 = \left\langle \dfrac{\mathit{freq}(y, Y_2)}{\mid Y_2 \mid}  \right\rangle_{y \in \mathcal{Y}} 
        \end{equation}
        Given $p_1$ and $p_2$, measuring impurity amounts to measuring their distance 
        to the standard basis of the class space: the larger the distance, the more impure the split.
        Conventionally, the distance is then mapped to a single a scalar through a weighted sum.
        Typically, Euclidean or cosine distance is adopted~\cite{DBLP:conf/aaai/FayyadI92}.
        A notable impurity measure is the Gini, namely the expected probability of incorrectly labeling a random instance:
        \begin{equation*}
            \mathit{Gini(n_i)} = 1 - \sum_{y \in \mathcal{Y}} \left(\dfrac{\mathit{freq}(y, Y_i)}{\mid Y_i \mid}\right)^2
        \end{equation*}
        The difference between the Gini of the parent node and the weighted average Gini of the child nodes quantifies the increase in impurity after a candidate split.
        Gini-based approaches, such as CART \cite{DBLP:books/wa/BreimanFOS84}, select the split which maximizes such an increase.

      \textit{Information Gain splits.} Entropy measures the information contained within a random variable based on the uncertainty of its events \cite{Cover1999ElementsIT}. The standard is \textit{Shannon's entropy} \cite{DBLP:conf/icaisc/MaszczykD08}, which for a node $n_i$ is defined as:
        \begin{equation*}
          \label{eq:ShannonEntropy}
          H(n_i) = \sum_{y \in \mathcal{Y}} - \dfrac{\mathit{freq}(y, Y_i)}{\mid Y_i \mid} \log \dfrac{\mathit{freq}(y, Y_i)}{\mid Y_i \mid}
        \end{equation*}
      Therefore, entropy is the expected information of the class distribution at the node $n_i$. The difference of entropy of the parent node and the weighted average entropy of the child nodes is the Information Gain of a split:
      \begin{equation*}
        IG = H(n_0) - \sum\limits_{i \in \{1,2\}} \dfrac{\mid Y_i \mid}{\mid Y_0 \mid} H(n_i)
      \end{equation*}
      Selecting the split for which the Information Gain is maximized is a popular strategy, which has been adopted in several learning algorithms, including ID3~\cite{id3} and GID3~\cite{gid3}, C4~\cite{c4} and C4.5~\cite{c4.5}.

      \textit{Statistical test splits.} Other strategies look for the best split through purely statistical tests or confidence intervals.
        Starting from the contingency tables of true and predicted class labels at the nodes $n_0$, $n_1$ and $n_2$, a number of statistical measures of associations can be considered to be maximized~\cite{chisquared,DBLP:conf/icdm/KatzSRO12,DBLP:conf/ijcnn/RosaC15}. 

        \subsection{Multivariate splits}
            Unlike univariate splits, the search space of parameters $\alpha_i$'s and $\beta_i$'s in (\ref{eq:splits}) for multivariate splits cannot be efficiently enumerated.
            Instead, they are determined through specialized optimization algorithms. 
            Such an optimization introduces a significant computational overhead.
          
            \textit{Margin optimization.} Given their success in a host of applications, Support Vector Machines~\cite{svm} (SVMs) are a natural candidate for optimizing class separation.
            Provided with linear kernels, SVM-based multivariate trees have been implemented in SVM Trees~\cite{svmtree}, Geometric Trees~\cite{geometrictree}, and in Reflector-based Trees such as CartOpt~\cite{cartopt} and HHCart~\cite{hhcart}.
            
            \textit{Linear optimization.} Linear Trees are families of DTs that directly optimize a linear model at each node.
            Linear Discriminant Trees~\cite{DBLP:journals/ijprai/YildizA05} and QUEST~\cite{loh1997split} build on top of Fisher's linear discriminant, a statistical method to find linear combinations separating samples.
            Linear Machines~\cite{linearmachines} relies on the encoding of a DT into a set of one-vs-all linear models.
            Weighted Oblique Decision Trees~\cite{weightedodt} look to optimize a differentiable formulation of entropy in which samples are first mapped through a nonlinear function.
            
            \textit{Composed optimization.} Trees based on composed optimization aim to leverage univariate splits to either learn mixed trees, or use the univariate split as an initialization for the multivariate one.
            For instance, OC1~\cite{oc1} initializes each multivariate split with a univariate one derived from CART, which is then replaced by a better multivariate one learned through a series of optimization techniques.
            Directly extending its univariate counterpart, CART-LC~\cite{DBLP:books/wa/BreimanFOS84} optimizes a local linear model, which is then simplified by pruning coefficients.
            Model Trees~\cite{modeltrees} instead take a mixed approach and first build a CART Tree, by progressively replacing the univariate splits at the penultimate layer with multivariate ones, each iteration pruning the replaced subtree, only to stop when no further performance gain is achieved.
            
            \textit{Other forms of optimization.} Several learning algorithms employ complex nonlinear and nondifferentiable optimization techniques: Genetic Trees~\cite{DBLP:conf/icaisc/Kretowski04} and Evolutionary Trees~\cite{DBLP:journals/tec/Cantu-PazK03} employ genetic programming, and Annealing Trees~\cite{DBLP:conf/ijcai/HeathKS93} employ simulated annealing.
            Another noteworthy family of approaches departs from a greedy search by encoding the tree learning into a global MILP optimization problem~\cite{DBLP:conf/cp/BessiereHO09,DBLP:journals/ml/BertsimasD17,DBLP:conf/cpaior/VerwerZ17,DBLP:conf/nips/ZhuMPNK20}. The resulting DTs are called optimal trees.
      %


    
    \subsection{Feature correlation}
    In this paper, we focus on feature correlation as a major property of interest due to its generality and possible impact on DT performances.
    Correlation, and even more so multicollinearity, relates multiple features by quantifying the linear relationship occurring among them.
    Generally, given two linearly dependent features $x_i$, $x_j$, we can express one in linear terms of the other, i.e.:
    \begin{equation}\nonumber
      x_i = \alpha x_j + \beta + \epsilon,
    \end{equation}
    for suitable $\alpha, \beta \in \mathbb{R}$ and small error $\epsilon$.
    The same relationship can be found in sets of $\Delta$ features $x_i, \dots, x_{i + \Delta}$, in which case we are dealing with \emph{multicollinearity}, that is:
    \begin{equation}\nonumber
      x_i = \sum_{j = i + 1}^\Delta \alpha_{j} x_{j} + \beta + \epsilon,
    \end{equation}
    again for suitable $\alpha_{i + 1}, \dots, \alpha_{i + \Delta}, \beta$ and small error $\epsilon$.
    Perfect multicollinearity is rare in practice, and sets of features are said to enjoy collinearity even if they are in an approximately linear relationship.
    %
    Multicollinearity leads to several   problems in regression analysis, which include high instability in the optimization algorithm itself and large variance in the expected results.
    As a consequence, removal of highly correlated, or approximately collinear, sets of features has been always a standard practice in dataset pre-processing \cite{Multicollinearity}.
    Standard metrics to measure collinearity, such as Variance Inflation Factors~\cite{belsley2005regression}, are tailored to specific cases, e.g., linear regression, hence dataset- and task-agnostic proxy metrics, such as correlation, are typically used instead.
    Correlation can be quantified in many forms, each trying to detect a slightly different relationship between pairs of features $(x_i, x_j)$.
    Pearson's correlation $\rho^P$ measures linear correlation through normalized covariance.
    %
    %
    Spearman's correlation $\rho^S$ instead aims to measure monotonic correlation  through ranks of features.
    %
    %
    Kendall's correlation $\rho^\tau$, also known as Kendall's $\tau$, measures concordance of order, that is, the probability of observing a difference between concordant and discordant pairs w.r.t. the orderings assigned by $x_i$ and $x_j$.
    
    %
    %
    %
    
    \begin{table}[t]
      \centering
      \begin{tabular}{ @{} l r r r @{} }
        \toprule
        & \textbf{Mean $\pm$ Stdev} & \textbf{Min} & \textbf{Max} \\
        \toprule
        \textbf{Dataset size}                       & 196,907 $\pm$ 1,301,496 & 68 & 9,999,889 \\
        \textbf{Dimensionality}             & 649.22 $\pm$ 2,942.14 & 4 & 20,001 \\
        \textbf{Dimensionality ratio}  & 20,585.62 $\pm$ 144,532.71 & 0.00 & 1,111,098.77 \\
	\bottomrule
      \end{tabular}
      \caption{Summary of the 57 benchmark datasets.}
      \label{tbl:datasets_summary}
    \end{table}
    %
    %
      %
      %
      \begin{figure*}[t!]
        \centering
        \begin{subfigure}[b]{0.45\textwidth}
            \includegraphics[width=\textwidth]{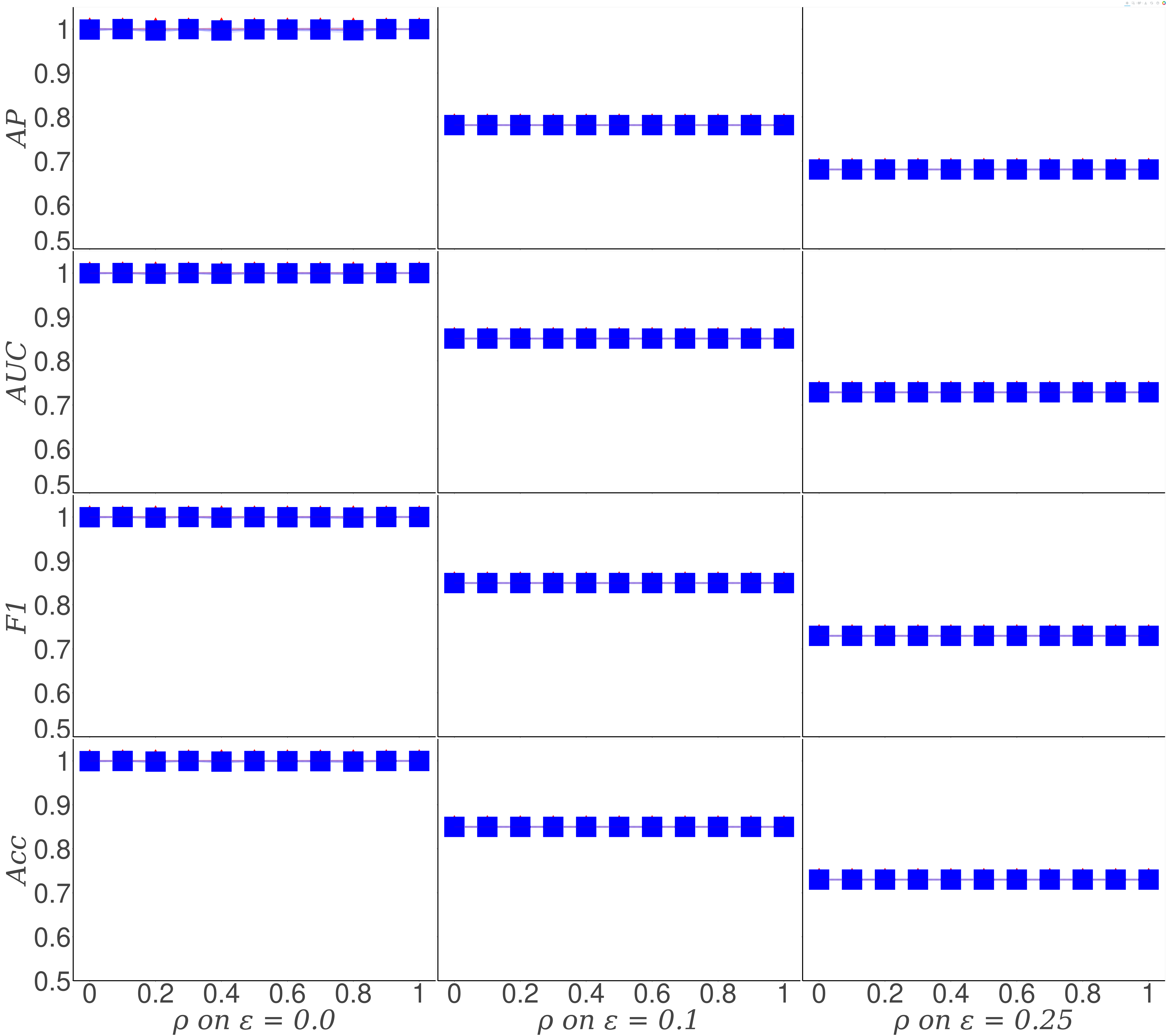}
            \caption{$\theta = 0\degree$}
        \end{subfigure}
        \begin{subfigure}[b]{0.45\textwidth}
            \includegraphics[width=\textwidth]{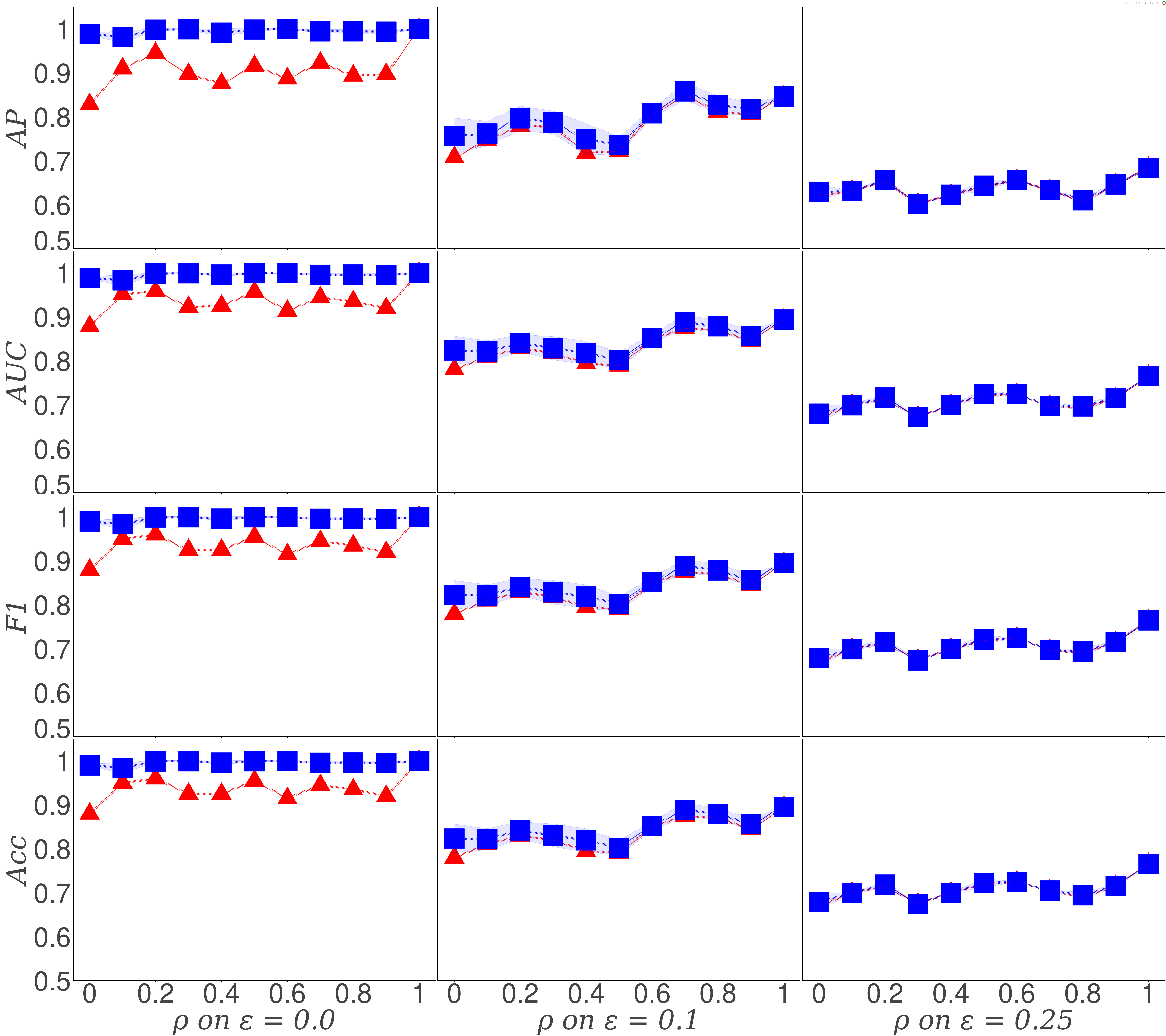}
            \caption{$\theta = 15\degree$}
        \end{subfigure}
        \begin{subfigure}[b]{0.45\textwidth}
            \includegraphics[width=\textwidth]{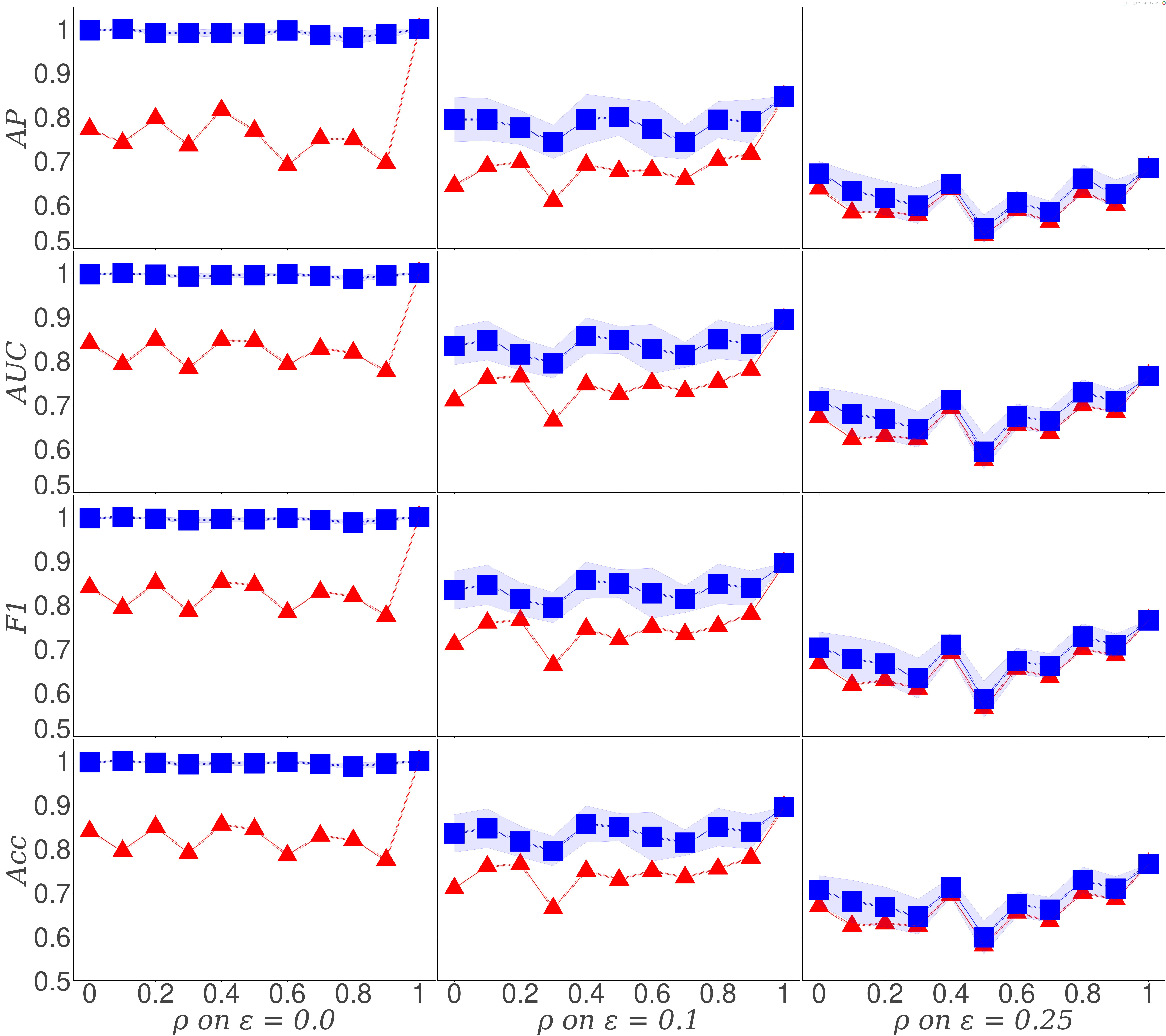}
            \caption{$\theta = 30\degree$}
          \end{subfigure}
          \begin{subfigure}[b]{0.45\textwidth}
            \includegraphics[width=\textwidth]{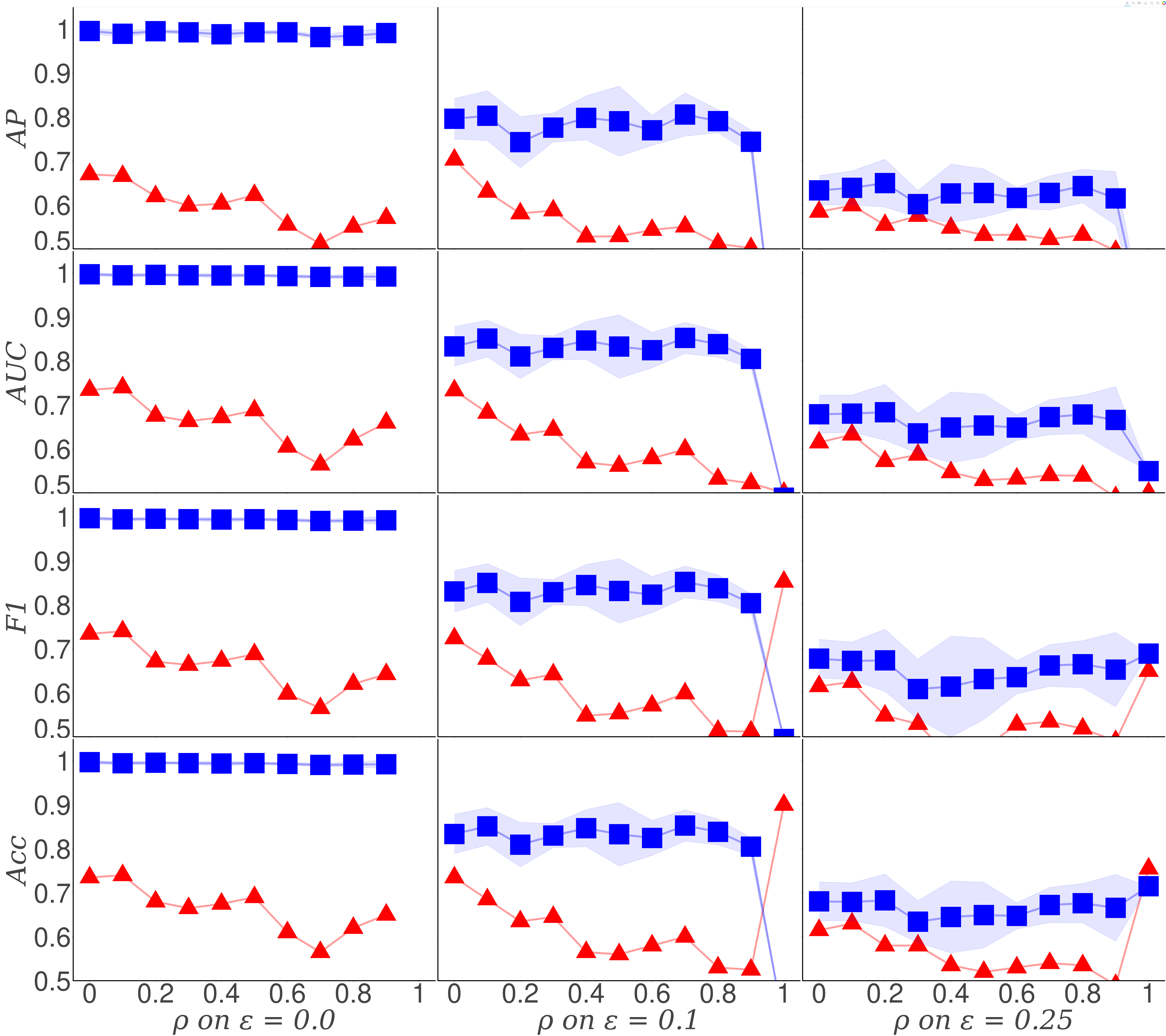}
            \caption{$\theta = 45\degree$}
          \end{subfigure}
          \caption{Performances of univariate and multivariate DTs  with maximum depth of 1 on synthetic datasets with slope angle $\theta$, correlation $\rho$ and noise $\epsilon$. Legend: red triangles for UDTs, blue squares for MDTs.}
          \label{fig:exp:synthetic_univariate_vs_multivariate_splits}
        \end{figure*}
      \begin{figure*}[t!]
        \centering
        \begin{subfigure}[b]{0.45\textwidth}
          \includegraphics[width=\textwidth]{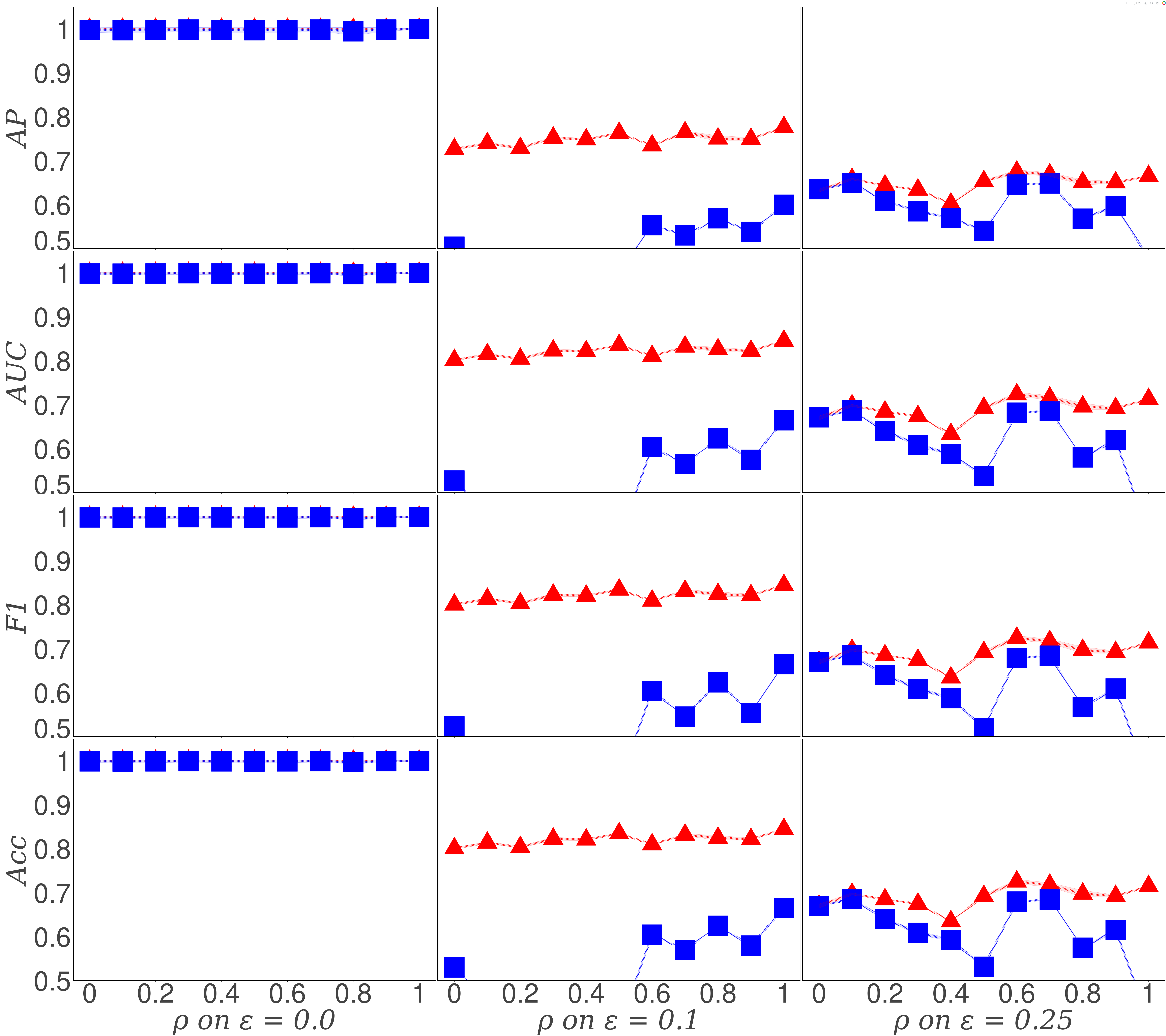}
            \caption{$\theta = 0\degree$}
        \end{subfigure}
        \begin{subfigure}[b]{0.45\textwidth}
            \includegraphics[width=\textwidth]{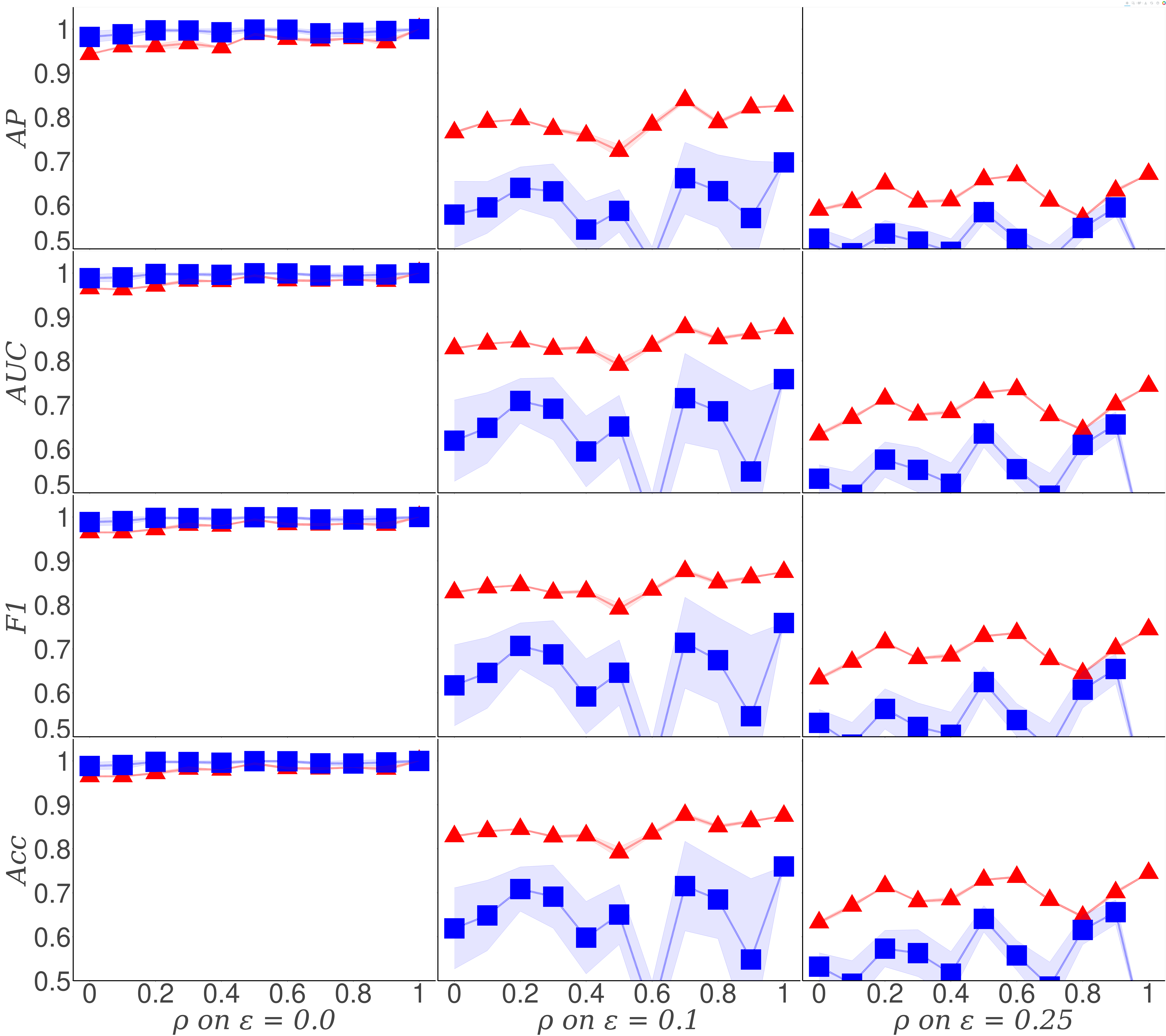}
            \caption{$\theta = 15\degree$}
          \end{subfigure}
          \begin{subfigure}[b]{0.45\textwidth}
            \includegraphics[width=\textwidth]{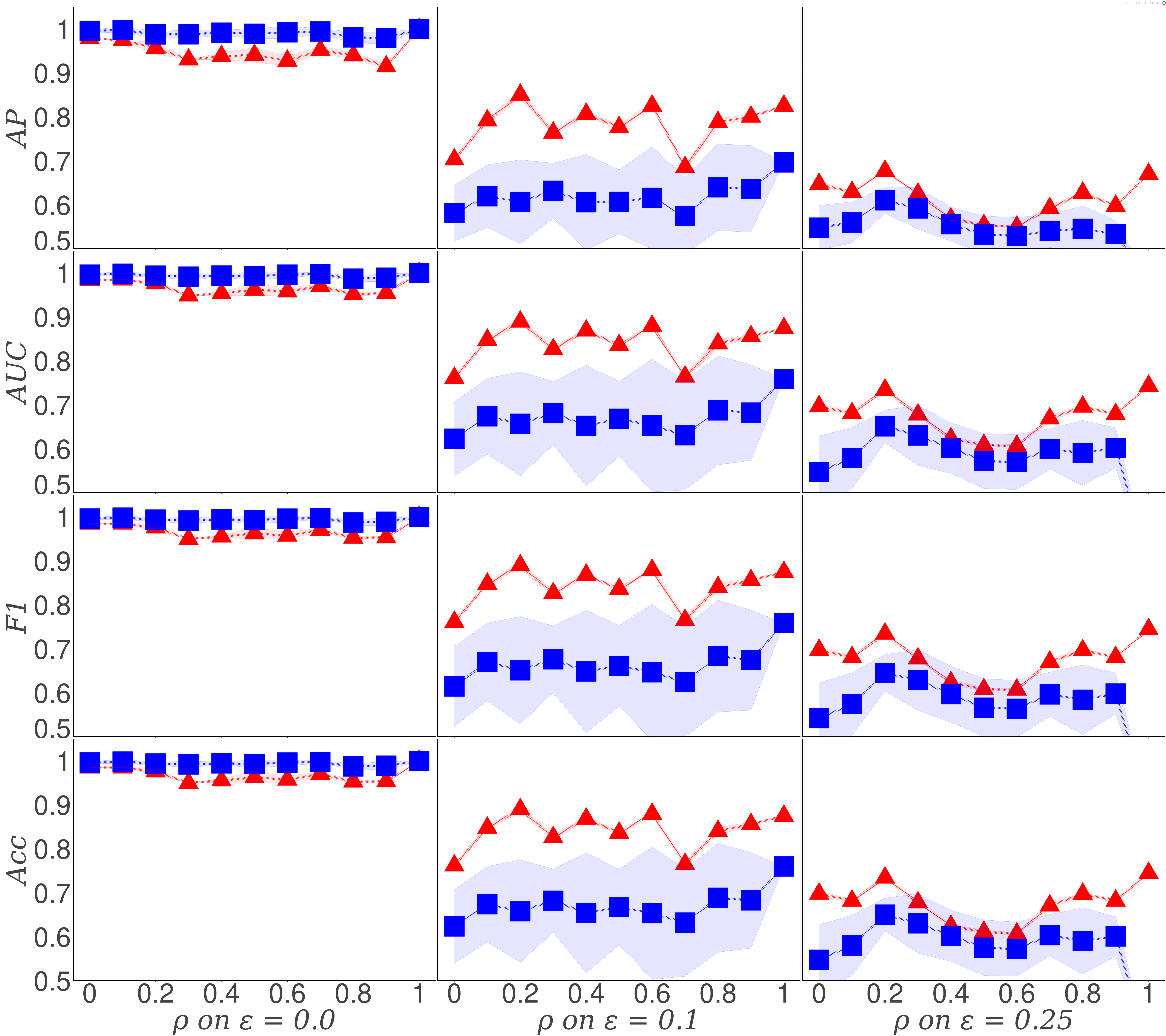}
            \caption{$\theta = 30\degree$}
          \end{subfigure}
          \begin{subfigure}[b]{0.45\textwidth}
            \includegraphics[width=\textwidth]{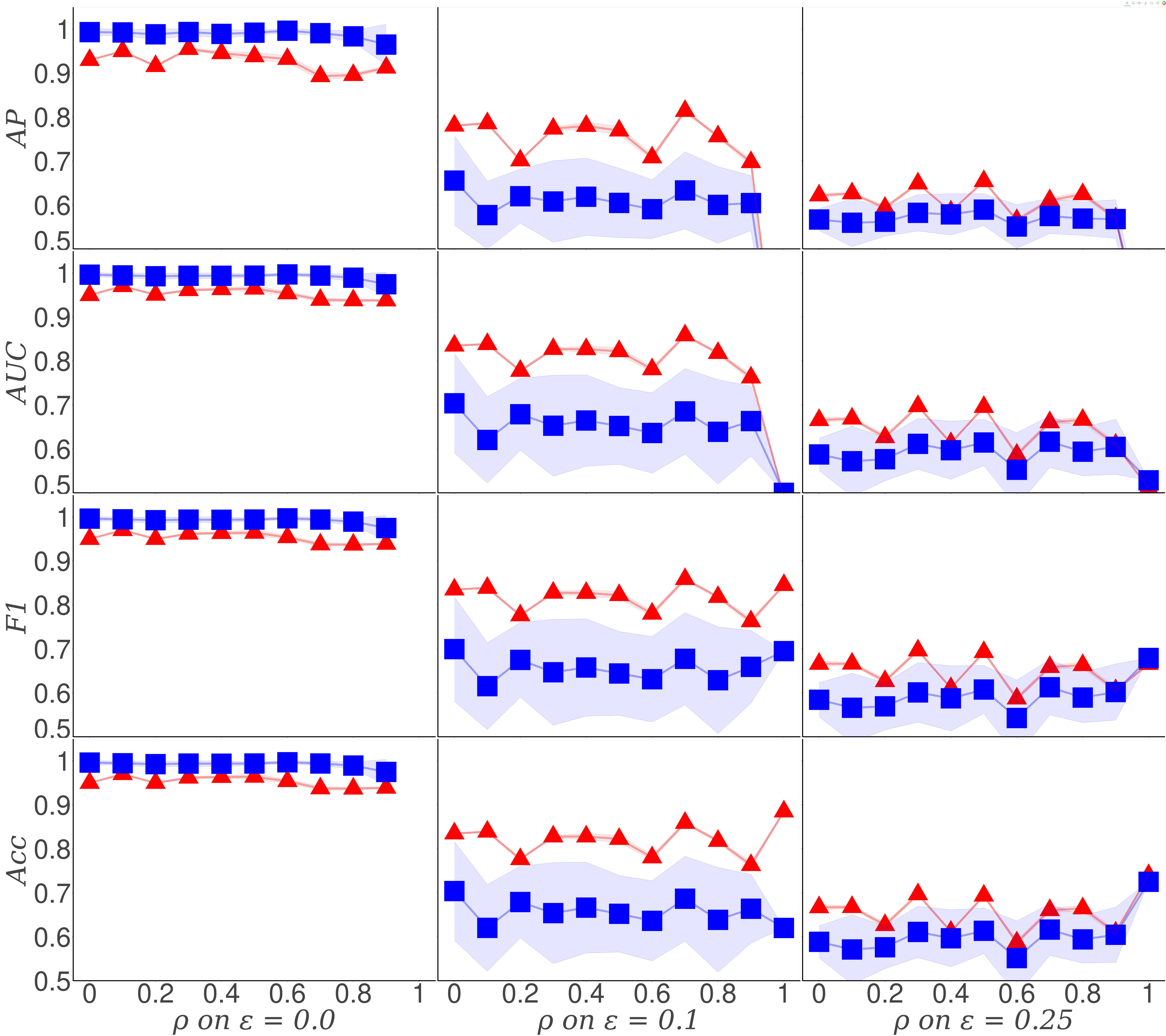}
            \caption{$\theta = 45\degree$}
        \end{subfigure}
        \caption{Performances of univariate and multivariate DTs with maximum depth of 16 on synthetic datasets with slope angle $\theta$, correlation $\rho$ and noise $\epsilon$. Legend: red triangles for UDTs, blue squares for MDTs.}
        \label{fig:exp:synthetic_univariate_vs_multivariate_trees}
      \end{figure*}

      \begin{table*}[t!]
        \centering
	      \begin{tabular}{@{} c r r c r r c r r c r r @{}}
		      \toprule
          \multirow{2.4}*{$\theta$} & \multicolumn{2}{c}{\textbf{Accuracy}} && \multicolumn{2}{c}{\textbf{F1}} && \multicolumn{2}{c}{\textbf{AUC}} && \multicolumn{2}{c}{\textbf{AP}} \\
            \cmidrule{2-3} \cmidrule{5-6} \cmidrule{8-9} \cmidrule{11-12} 
            & \textbf{Test outcome} & $p$ \textbf{value} && \textbf{Test outcome} & $p$ \textbf{value} && \textbf{Test outcome} & $p$ \textbf{value} && \textbf{Test outcome} & $p$ \textbf{value} \\
            \midrule$0\degree$	& $0.005$ 	& $10^{-1}$	&& $0.005$	& $10^{-1}$ 	&& $0.005$ 	& $10^{-1}$ 	&& $0.005$	& $10^{-1}$ \\ 
            $15\degree$ 	& $-0.136$	& $10^{-28}$ 	&& $-0.136$ 	& $10^{-28}$	&& $-0.132$ 	& $10^{-27}$	&& $-0.144$ 	& $10^{-30}$ \\ 
            $30\degree$ 	& $-0.159$	& $10^{-32}$ 	&& $-0.159$ 	& $10^{-32}$	&& $-0.158$ 	& $10^{-32}$	&& $-0.159$ 	& $10^{-32}$ \\ 
            $45\degree$ 	& $-0.270$	& $10^{-68}$ 	&& $-0.270$ 	& $10^{-68}$	&& $-0.273$ 	& $10^{-69}$	&& $-0.246$ 	& $10^{-59}$ \\ 
            \bottomrule
        \end{tabular}
        \caption{Paired t-test of the performance of univariate and multivariate DTs with maximum depth of 16 on synthetic datasets with $\epsilon=0$ and slop angle $\theta$.}
        \label{tbl:exp:t_test_no_noise}
    \end{table*}

      \begin{figure*}[t!]
     \vspace{5ex}
        \centering        \includegraphics[width=0.5\textwidth]{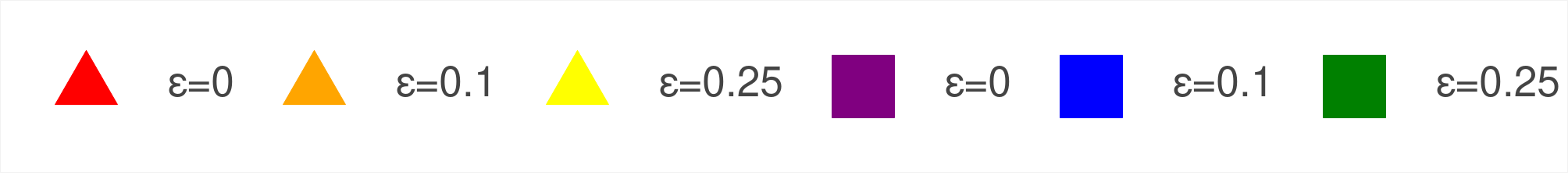}
        \hfill
        \\
        \includegraphics[width=\textwidth]{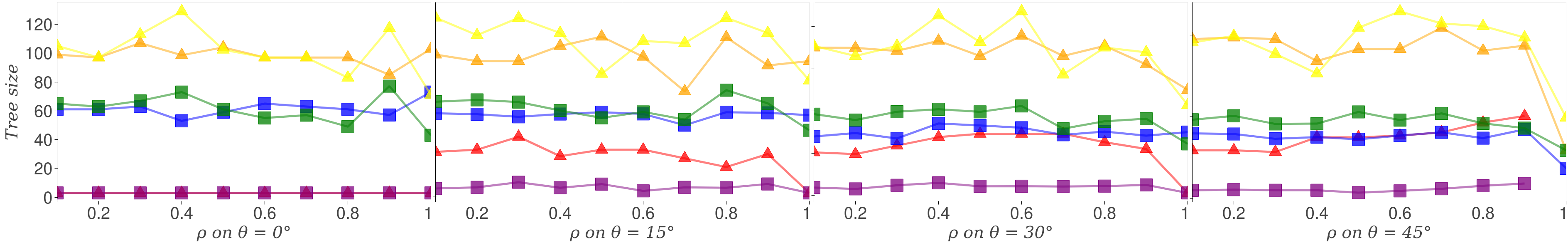}
        \caption{Model complexity of  univariate and multivariate DTs with maximum depth of 16 on synthetic datasets with slope angle $\theta$, correlation $\rho$ and noise $\epsilon$.  Legend:  triangles for UDTs, squares for MDTs.}        \label{fig:exp:synthetic_univariate_vs_multivariate_size}
      \end{figure*}
        \begin{figure*}[t!]
          \centering
          \begin{subfigure}[b]{0.32\textwidth}
            \centering
            \includegraphics[width=\textwidth,trim={0 1.45cm 0 0},clip]{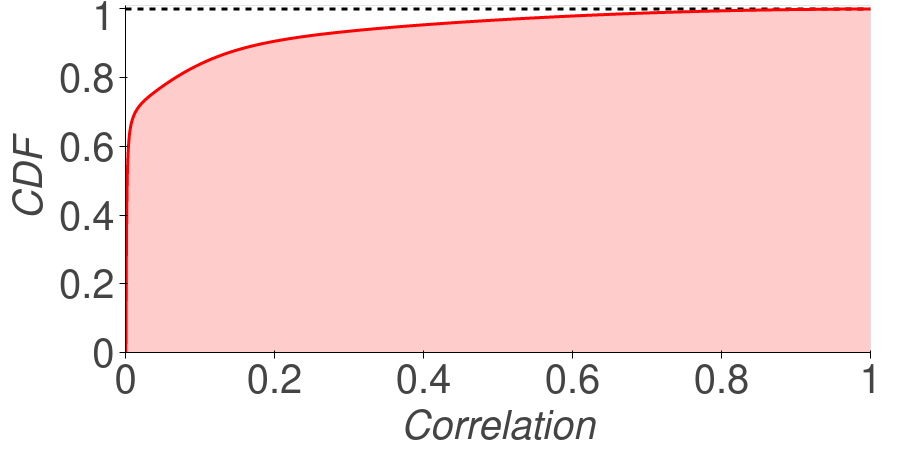}
          \end{subfigure}
          \begin{subfigure}[b]{0.32\textwidth}
            \centering
            \includegraphics[width=\textwidth,trim={0 1.45cm 0 0},clip]{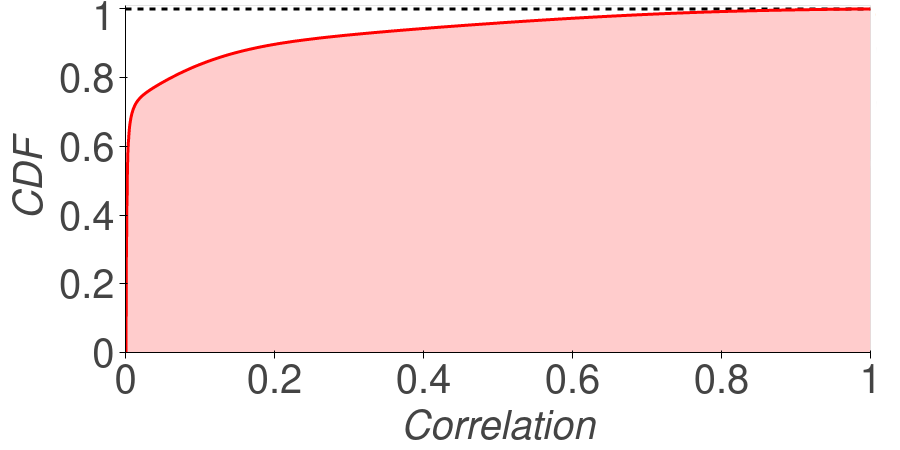}
          \end{subfigure}
          \begin{subfigure}[b]{0.32\textwidth}
            \centering
            \includegraphics[width=\textwidth,trim={0 1.45cm 0 0},clip]{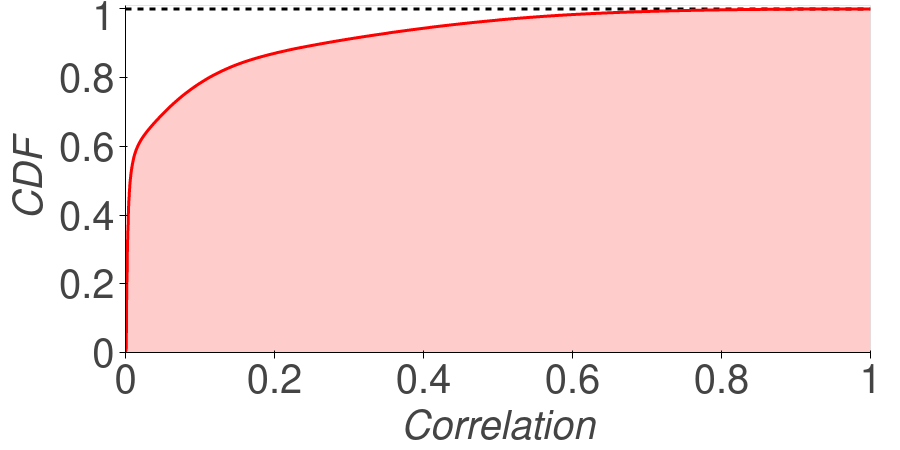}
          \end{subfigure}
          \hfill
          \begin{subfigure}[b]{0.32\textwidth}
            \centering
            \includegraphics[width=\textwidth]{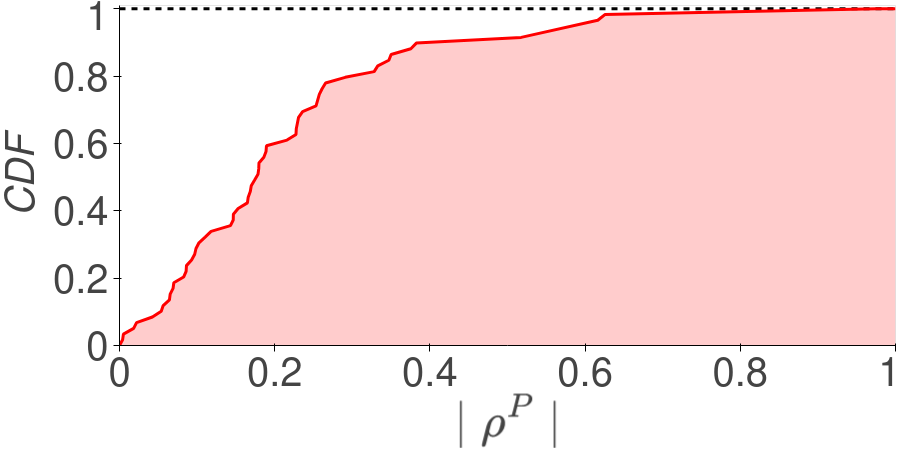}
            \caption{Pearson}
          \end{subfigure}
          \begin{subfigure}[b]{0.32\textwidth}
            \centering
            \includegraphics[width=\textwidth]{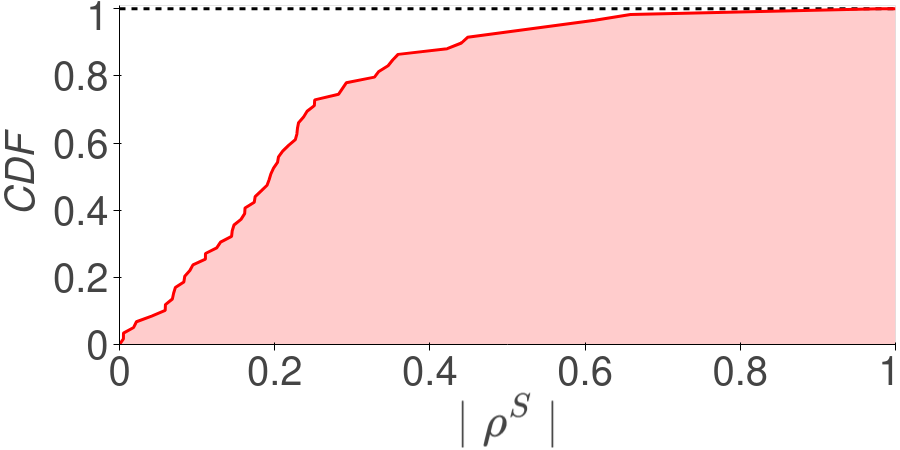}
            \caption{Spearman.}
          \end{subfigure}
          \begin{subfigure}[b]{0.32\textwidth}
            \centering
            \includegraphics[width=\textwidth]{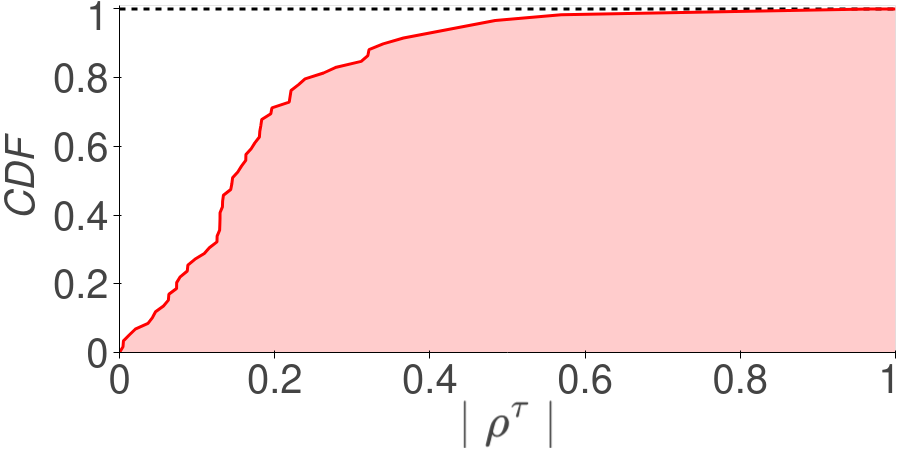}
            \caption{Kendall.}
          \end{subfigure}
          \hfill
          \caption{Empirical CDFs of absolute feature correlation over the benchmark datasets. Top: any pair of features in a same dataset. Bottom: mean correlation per dataset.}
          \label{fig:rq2}
        \end{figure*}
        %

    \section{Research Questions}
    \label{sec:method}

      Regarding univariate splits, \cite{DBLP:journals/ml/Martin97} observes that the ``choice of split-selection metric typically has little effect on accuracy, but can profoundly affect complexity and the effectiveness and efficiency of pruning".
      Regarding Linear Discriminant Trees, while ``the proposed method is accurate, learns fast, and generates small trees, [...] results indicate that in the majority of cases, a univariate method suffices" \cite{DBLP:journals/ijprai/YildizA05}.
    
      The reasons for the above conclusions have not been sufficiently clarified in the (forty years old) literature on DT learning. We intend to answer the following  questions.

    \begin{description}
      \item[RQ1.] \emph{Do feature correlation or other factors, such as complexity of decision boundary and label noise, impact on the performances of the split functions?}
      In particular, on the relative strength of univariate vs multivariate splits?
      By answering this research question we aim to understand the relationship between the correlation among the features and the UDTs/MDTs performances.
      Detecting such an impact will allow to guide practitioners on the choice of UDTs vs MDTs.

      \item[RQ2.] \emph{Are standard benchmark datasets used for evaluating DT learning algorithms biased?}
      We analyze standard benchmark datasets that have been and/or are currently used in the literature.
      Identifying common patterns of such datasets, i.e., with regard to feature correlation, will then allow us to understand whether experiments in the literature have relied on biased collections of datasets.

      \item[RQ3.] \emph{Does the bias in benchmark datasets transfer to a biased evaluation of the performances of DT learning algorithms?}
      On the basis of the answers to RQ1 and RQ2, we want to understand if bias in benchmark datasets plus dependence of split functions from correlation of features and other factors, turn out to impact the experimental results in favor of UDTs vs MDTs.
      
    \end{description}

    \section{Experiments}
    \label{sec:exp}

    \subsection{Benchmark datasets}
    \label{sec:datasets}
    Following~\cite{DBLP:journals/jmlr/DelgadoCBA14}, we base the bulk of our analysis on a large subset of UCI\footnote{\href{https://archive-beta.ics.uci.edu}{https://archive-beta.ics.uci.edu}} datasets.
    Moreover, we include several datasets from the OpenML repository\footnote{\href{https://www.openml.org}{https://www.openml.org}}, and from the Kaggle platform\footnote{\href{https://www.kaggle.com}{https://www.kaggle.com}}, for a total of 57 datasets\footnote{Several datasets from the original paper~\cite{DBLP:journals/jmlr/DelgadoCBA14} have been removed due to their extremely low number of features.}.
    We present a full list of datasets in Table~\ref{tbl:apdx:datasets} in the Appendix, and a short summary in Table~\ref{tbl:datasets_summary}.
    Datasets vary wildly in size, ranging from $\approx$ 70
    to $\approx$ 10M instances, in dimensionality, ranging from 4 to 20,000 features, and in the size to dimensionality ratio, ranging from 0.009 to 110,000. We left in datasets only the numeric attributes, since we restrict to linear split functions of the form (\ref{eq:splits}). Finally, feature values are z-score normalized.

    \subsection{Synthetic datasets}
    \label{sec:synthetic_dataset}
    To best estimate the effects of feature correlation and label noise on DT learning algorithms, we experiment also with synthetic data in which we control for these factors.
    We generate datasets with two normally-distributed
    features, $X_1$ and $X_2$, including $1,000$ instances, and with Pearson's correlation\footnote{$X_1$ and $X_2$ are generated as follows. Starting from $X_1, Z \sim N(0, 1)$, we have that $X_2 = \rho \cdot X_1 + \sqrt{1-\rho^2}  \cdot Z \sim N(0, 1)$ and $Cor(X_1, X_2) = \rho$.} $\rho^P = \rho$ ranging from $0$ to $1$ in $0.1$ steps.
    The binary class label is defined as $Y = \mathds{1}(X_2 > m \cdot X_1)$, where the parameter $m = tan(\theta)$ is the slope of the decision boundary and $\theta$ is the slope angle.
    Moreover, each dataset is further replicated and perturbed to introduce some noise by randomly flipping each label according to the outcome of a Bernoulli trial with parameter $\epsilon \in \{0, 0.1, 0.25\}$. The definition of the class feature $Y$ is specifically intended to distinguish the cases when multivariate splits are theoretically the best solution ($0\degree \ll \theta \ll 90\degree$) from cases where axis-parallel splits are sufficient ($\theta \approx 0\degree$ or $\theta \approx 90\degree$) . Experiments will then show the impact of feature correlation and label noise in such  contexts.

    \subsection{DT learning algorithms}
    \label{sec:models}
    For univariate DTs, we use CART~\cite{DBLP:books/wa/BreimanFOS84}, while for multivariate DTs we implemented an Omnivariate Tree~\cite{DBLP:journals/ijprai/YildizA05}.
    Our implementation tests several splits at each node, namely: an SVM split~\cite{svmtree}, a gradient-SVM split, a Ridge split, a Least Squares split, an Elastic Net split~\cite{friedman2010regularization}, a Lasso split, and a CART split.
    At each node, all splits are evaluated, and the one yielding the best Gini is selected.
    As in composed optimization DTs such as OC1~\cite{oc1}, we include a CART split in the candidates in case the data can be best separated with a univariate split.
    Due to their extremely large computational cost on moderate-to-large datasets, we do not include Optimal Trees in the experimentation.

    \subsection{Performance and model complexity metrics}
    Each synthetic and benchmark dataset is split into $90\%$ training set and $10\%$ test set by a stratified hold-out method. 
    We evaluate the learned DTs through several performance metrics on the test set, including: F-measure (F1), AUC-ROC (AUC), Average Precision (AP), and Accuracy (Acc).
    As for model complexity, we consider: the size of the DT, and, for MDTs, also the fraction of non-zero coefficients at its nodes.

   \subsection{Experimental Results}
   
   Let us consider our research questions.
   
    \subsection*{RQ1. Do feature correlation, decision boundary slope, and label noise impact on the performances of the split functions?}
      \label{sec:tree_properties}

Figure~\ref{fig:exp:synthetic_univariate_vs_multivariate_splits} shows the mean and standard deviation of the performances of DTs with a single univariate or multivariate split. Each quadrant referes to synthetic datasets with increasing  slope angle $\theta \in \{0\degree, 15\degree, 30\degree, 45\degree\}$. 
Let's first consider the case of no noise, i.e., $\epsilon =  0$. By construction of the synthetic datasets, angles close to $0\degree$ should lead to axis-parallel splits, while angles close to $45\degree$ should favor oblique splits. This is apparent in the plots, where for $\theta = 0\degree$ there is no difference between UDTs and MDTs, and for $\theta=45\degree$ the MDTs perform much better than UDTs.  Moreover, for $\theta = 45\degree$, the gap between MDTs and UDTs increases with correlation $\rho$. This can be explained by observing that the larger the $\rho$ the more the data instances are closer to the decision boundary $Y = \mathds{1}(X_2 > X_1)$ of $\theta = 45\degree$, hence it becomes more and more difficult to separate them with axis-parallel splits.
For moderate ($\epsilon=0.1$) to large ($\epsilon=0.25$) noise, the gap between MDTs and UDTs decreases. Intuitively, the decision boundary becomes more complex, and neither an oblique nor an axis-parallel split are adequate.

Figure~\ref{fig:exp:synthetic_univariate_vs_multivariate_trees} reports the same analysis for DTs of maximum depth $16$. For zero noise, the performance gap is smaller than in the previous case, as UDTs are now larger and can better separate the decision boundary.
However, Table~\ref{tbl:exp:t_test_no_noise} shows that such differences are statistically significant. For moderate noise ($\epsilon=0.1$), something surprising can be observed: UDTs perform better than MDTs, which in addition appear to be unstable. We explain this by the fact that MDTs overfit the decision boundary with unnecessarily complex oblique splits. This is particularly striking for $\theta=0\degree$, where axis-parallel splits are enough. For large noise ($\epsilon=0.25$), the performances of UDTs are still better than those of MDTs, but the gap is smaller -- as the irregularity of the decision boundary makes both axis-parallel and oblique splits inadequate.
Regarding model complexity, Figure~\ref{fig:exp:synthetic_univariate_vs_multivariate_size} shows that, regardless of label noise, UDTs are consistently larger than MDTs, with up to 50\% additional nodes.

\textit{In summary}, the presence of noise appears to have a major impact in favor of UDTs performances, but at the cost of higher model complexity. In absence of noise, MDTs are better than UDTs if the decision boundary is actually oblique. In such a case, the larger the correlation between features the larger the gap with UDTs.

    %
   
        \begin{table}[t]
          \centering
          \begin{tabular}{ @{} r@{\hspace{2ex}}r@{\hspace{2ex}}r@{\hspace{2ex}}r@{\hspace{2ex}}r @{} }
            \toprule
            \textcolor{white}{MDTs} & \multicolumn{1}{c}{\textbf{Accuracy}} & \multicolumn{1}{c}{\textbf{F1}} & \multicolumn{1}{c}{\textbf{AUC}} & \multicolumn{1}{c}{\textbf{AP}}\\
            \midrule
            \multicolumn{4}{l}{\text{UDTs}} \\
            TR & $0.932 \pm 0.091$ & $0.928 \pm 0.098$ & $0.886 \pm 0.133$ & $0.818 \pm 0.230$  \\
            TS  & $0.844 \pm 0.125$ & $0.837 \pm 0.134$ & $0.771 \pm 0.163$ & $0.635 \pm 0.281$ \\
            \midrule
            \multicolumn{4}{l}{\text{MDTs}} \\
            TR  & $0.811 \pm 0.213$ & $0.816 \pm 0.211$ & $0.785 \pm 0.203$ & $0.687 \pm 0.312$ \\
            TS & $0.720 \pm 0.207$ & $0.720 \pm 0.215$ & $0.683 \pm 0.195$ & $0.558 \pm 0.296$ \\
            \bottomrule
          \end{tabular}
          \caption{Performances of UDTs and MDTs on benchmark datasets. TR = training set. TS = test  set.}
          \label{tbl:rq3:absolute}
        \end{table}

        \begin{table}[t!]
        	\centering
        	\begin{tabular}{ @{}crrrr@{} }
        		\toprule
        		\textcolor{white}{TR} & \multicolumn{1}{c}{\textbf{Accuracy}} & \multicolumn{1}{c}{\textbf{F1}} & \multicolumn{1}{c}{\textbf{AUC}} & \multicolumn{1}{c}{\textbf{AP}}\\
        		\midrule
        		\text{mean}    & $.122 \pm .170$  & $.116 \pm .169$  & $.900 \pm .148$ & $.794 \pm .135$  \\
        		\text{min}    & $-.143$   & $-.212$  & $-.196$ & $-.302$\\
        		\text{max}    & $.681$   & $.770$   & $.642$  & $.486$\\
        		\text{UDT wins}   & $49$  & $47$  & $46$   & $46$\\
        		\text{MDT wins}   & $8$    & $10$  & $11$   & $11$\\
        		\bottomrule
        	\end{tabular}
        	\caption{Performance gap between UDTs and MDTs on benchmark datasets (test set).}
        	\label{tbl:rq3:difference}
        \end{table}
        
        \begin{table}[t!]
          \centering
          \begin{tabular}{ @{} r r r r @{} }
            \toprule
            \textcolor{white}{Multivariate} & \textbf{Tree size} & \textbf{Non-zero coefficients ratio}\\
            \midrule
            UDTs  & 57.00 $\pm$ 146.719    &  -  \\
            MDTs  & 59.00 $\pm$ 75.544  &    0.409 $\pm$ 0.241  \\
            \bottomrule
          \end{tabular}
          \caption{Model complexity of UDTs and MDTs on benchmark datasets.}
          \label{tbl:rq3:complexity}
        \end{table}
        \begin{figure}[t!]
            \centering
            \includegraphics[width=0.45\textwidth]{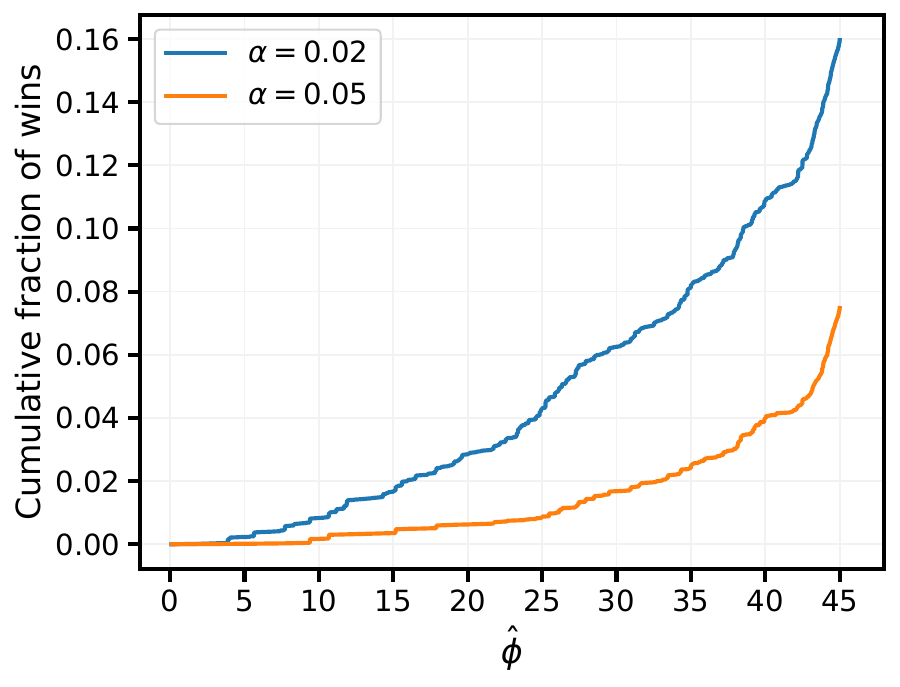}
            \caption{Cumulative weighted fraction of multivariate splits wins in benchmark datasets at the variation of slope angle estimate $\hat{\phi}$ and threshold $\alpha$.}
            \label{fig:slopes}
        \end{figure}
        %
        

        %
        \begin{table}[t!]
            \centering
            \begin{tabular}{ @{} c r c r r @{} }
            \toprule
            \textbf{Partition} &
            \textbf{Size} & & \textbf{Correlation} &  \multicolumn{1}{c}{$\hat{\phi}$}\\
            \midrule
            \multirow[c]{3}*{MDT $\gg$ UDT}  
            & \multirow[c]{3}*{$4$}  
            & $|\rho^\tau|$ & $0.448 \pm 0.132$   & \multirow[c]{3}*{$29.66\degree \pm 13.61\degree$} \\
                                                & & $|\rho^P|$ & $0.310 \pm 0.391$      & \\
                                                & & $|\rho^S|$ & $0.449 \pm 0.247$      & \\
            \midrule
            \multirow[c]{3}*{UDT $\gg$ MDT}  
            & \multirow[c]{3}*{$8$}      
            & $|\rho^\tau|$ & $0.035 \pm 0.123$   &  \multirow[c]{3}*{$21.81\degree \pm 13.44\degree$} \\ 
                                                & & $|\rho^P|$ & $0.200 \pm 0.306$      &  \\
                                                & & $|\rho^S|$ & $0.148 \pm 0.167$      &  \\
            \midrule
            \multirow[c]{3}*{MDT $\approx$ UDT}  
            & \multirow[c]{3}*{$45$}  
            & $|\rho^\tau|$ & $0.188 \pm 0.166$  & \multirow[c]{3}*{$22.06\degree \pm 14.05\degree$}  \\
                                                & & $|\rho^P|$ & $0.218 \pm 0.183$     &   \\
                                                & & $|\rho^S|$ & $0.218 \pm 0.183$     &   \\
            \bottomrule
            \end{tabular}
            
            \caption{Correlation and  $\hat{\phi}$ estimate in benchmark datasets, by group.}
            \label{tbl:rq3:correlations}
        \end{table}

    \subsection*{RQ2. Are standard benchmark datasets used for evaluating decision tree algorithms biased?}
        \label{sec:rq2}

        Figure~\ref{fig:rq2} shows the empirical cumulative distributions (CDFs) of  Pearson, Spearman, and Kendal correlation coefficients over the benchmark datasets. The top CDFs regard the correlation of any pair of features in the same dataset. The bottom CDFs regard the mean correlation at a dataset level.
        The feature correlation is extremely low over all pairs of features, as shown by the steep CDF curves at the top.
        Small mean correlation is also apparent in the CDFs at dataset level, with half of them having a correlation lower or equal than 0.2.
        Such a bias in favor of low correlated features can be traced back to the long-standing issue of collinearity in linear models,
        for which inference algorithms are able to converge faster and to less biased solutions.
        
        According to the experiments on synthetic data, the other factors affecting performance regard the decision boundary (the slope angle $\theta$) and the noise ($\epsilon$). However, we do not have ground truth knowledge to measure them over the benchmark datasets. We proceed with a (weak) approximation as follows. Consider any distinct pair $x_1$, $x_2$ of features in a dataset\footnote{For the high dimensional datasets \textit{arcene}, \textit{dexter}, and \textit{gisette}, we considered random samples of $1$M, $500$K and $100$K pairs respectively.}. We test whether the decision boundary w.r.t. these two-dimensional features can be better described by an univariate or a multivariate split. To this end, we train three Linear SVMs using as features: (1) $x_1$ only, (2) $x_2$ only, and (3) both $x_1$ and $x_2$. We say that the multivariate split from (3) \textit{wins} over the univariate splits in (1, 2) if the accuracy\footnote{Since we are not building predictive models in this task, the whole dataset is used both for training and for calculating the accuracy.} of the Linear SVM using (3) is greater than $1+\alpha$ times the accuracy of both Linear SVMs using (1) or (2). Here, $\alpha$ is a threshold to ensure that the improvement in accuracy is substantial. We experiment with $\alpha=0.02$ (two percent) and $\alpha=0.05$ (five percent). The fraction of pairs of features for which the multivariate split wins -- or simply, the \textit{fraction of wins} -- is a raw metric of degree of ``obliqueness" of the decision boundary of the dataset. Moreover,  
        we estimate the slope angle of the Linear SVM in (3) as $\hat{\theta} = tan^{-1}(-\beta_1/\beta_2)$, where $\beta_1, \beta_2$ are the  coefficients of $x_1$ and $x_2$ respectively. This approximation is weak if the unknown noise ($\epsilon$) is not small, or if the true decision boundary is not linear in $x_1$, $x_2$. 
        We perform the following transformation that makes the slope angle estimate independent of the order of the two features $x_1$ and $x_2$ and of the sign of the angle:
        \[ \hat{\phi} = \min\{\ |\hat{\theta}|,\ 90\degree - |\hat{\theta}|\ \} \]
        Thus, $\hat{\phi} \in [0\degree, 45\degree]$. 
        Figure~\ref{fig:slopes} shows the cumulative weighed\footnote{Each pair of features is weighted by the inverse of the total number of pairs in the dataset. This is made necessary by the large difference in feature dimensionality among the datasets.} fraction of wins at the variation of $\hat{\phi}$ for different thresholds $\alpha$. Remarkably, multivariate splits win over univariate splits with an improvement of at least $2$\% in less than $16$\% of the pairs, and with an improvement of at least $5$\% in less than $8$\% of the pairs. Therefore, decision boundaries of the benchmark datasets are mostly axis-parallel. Moreover, the skewed distributions in Figure~\ref{fig:slopes} also highlight that multivariate splits win mostly for large values of the slope angle $\hat{\phi}$. 
        
        
        \textit{In summary,} the benchmarks datasets exhibit  distributions of feature  correlation and of decision boundary slope that are skewed towards low correlation values and approximatively axis-parallel decision boundaries.

        
        %
    
    \subsection*{RQ3. Does the bias in benchmark datasets transfer to a biased evaluation of the performance of DT learning algorithms?}
        \label{sec:rq3}

        Table~\ref{tbl:rq3:absolute} reports the aggregate predictive performances of univariate and multivariate DTs on the benchmark datasets.
        Univariate DTs consistently achieves better Accuracy, F1, Average Precision, and AUC, as also shown in Table~\ref{tbl:rq3:difference}, alongside the number of wins per model.
        Interestingly, univariate DTs are better also on the training set, which suggests that the decision boundaries in the datasets lead multivariate DTs to largely overfit the data.
        The complexity of the learnt models, shown in Table~\ref{tbl:rq3:complexity}, is in favor of multivariate DTs, especially for what regards variability of tree size. However, the ratio of non-zero coefficients is, on average, quite large -- and this may limit the benefit of having smaller trees if the objective is to achieve model comprehensibility.
        The vast majority of datasets have performance gaps between UDTs and MDTs within one standard deviation from the mean, for each of the considered metrics.
        %
        Thus, we characterize a partition of datasets into three groups: when MDTs perform better than UDTs by at least one standard deviation (MDT $\gg$ UDT) for the F1 metric, the opposite case (UDT $\gg$ MDT), and when the performances are within one standard deviation (UDT $\approx$ MDT).
        Table~\ref{tbl:rq3:correlations} reports the mean $\pm$ stdev of the correlation coefficients, and of the slope angle estimate $\hat{\phi}$ over the pairs of features of the datasets in each group. For the small group MDT $\gg$ UDT, correlation is moderate-to-large and $\hat{\phi}$ is larger than for the other two groups. This is consistent with the results of experiments with synthetic data, where MDTs perform better than UDTs for large slope angles and large correlations.
        For the other two groups UDT $\approx$ MDT and UDT $\gg$ MDT, correlation is low and the estimated slop angles is, on average, lower than for the MDT $\gg$ UDT group. Again, this is in line with the case when oblique splits do not outperform axis-paralell splits. There is no clear difference between the groups UDT $\approx$ MDT and UDT $\gg$ MDT in terms of slope angle, while correlations coefficients of UDT $\gg$ MDT are smaller than those of UDT $\approx$ MDT. From our theoretical analysis, we can relate the different performance gaps of those two groups partly to different feature correlation and partly to different degrees of label noise. Unfortunately, without ground truth on the class attributes, we cannot test the latter hypothesis.
        
        \textit{In summary,} the observed performances of UDTs and MDTs can be largely explained by the factors analyzed in this paper: feature correlation, slope of the decision boundary, and, we conjecture, label noise. Based on the fact that benchmark datasets are skewed towards low correlation and axis-parallel decision boundaries, the fact that the observed performances are mostly in favor of UDTs are expected. The practitioner, then, should be warned about making the general conclusion that UDTs are always better than MDTs.

    \section{Conclusions}
        \label{sec:conclusions}
    
        We have compared two families of DTs which differ as per expressive power of the split function: univariate and multivariate DTs. The latter generates smaller trees, and, in absence of noise, performs better than univariate DTs. Moreover, the larger the feature correlation the larger is the improvement in performance. This was established in answering RQ1. Next, we observed that standard benchmark datasets are preprocessed to remove correlation. Also, we found their decision boundary is approximatively ``axis-parallel".
        This was established in answering RQ2.        
         Finally, we tested whether the previous two answers lead to the conclusion that the better performances of univariate DTs observed in the literature is due to biases in the benchmark datasets, namely skewed correlation and ``axis-parallel" decision boundaries. Our results support that the observed differences in performances can be explained by such factors. Consequently, practitioners are advised to test such factors on their datasets before making a choice whether to use MDTs or UDTs.
    

\section*{Datasets and code}
The benchmark datasets can be downloaded from \href{https://huggingface.co/mstz}{https://huggingface.co/mstz}.
Experimental Python code and scripts are available at \href{https://github.com/msetzu/Univariate-vs-multivariate-decision-trees}{https://github.com/msetzu/Univariate-vs-multivariate-decision-trees}.
    
\section*{Acknowledgment}

Research partly funded by PNRR - M4C2 - Investimento 1.3, Partenariato Esteso PE00000013 - ``FAIR - Future Artificial Intelligence Research" - Spoke 1 ``Human-centered AI", funded by the European Commission under the NextGeneration EU programme, and by the Excellent Science European Research Council (ERC) programme for the XAI project - ``Science and technology for the explanation of AI decision making" (g.a. No. 834756). 
This work reflects only the authors’ views and the European Research Executive Agency (REA) is not responsible for any use that may be made of the information it contains.

\bibliographystyle{IEEEtran}
\bibliography{main}

\appendix
    \begin{table*}
    \centering
    \begin{tabular}{@{} l r r  r | r c r | r c r | r c r | r c r | r @{}}
        \multirow{2.4}{*}{\textbf{Dataset}} & \multirow{2.4}{*}{\textbf{\# instances}} & \multirow{2.4}{*}{\textbf{\# features}} & \multicolumn{2}{c}{\textbf{Accuracy}} && \multicolumn{2}{c}{\textbf{F1}} && \multicolumn{2}{c}{\textbf{AUC}} && \multicolumn{2}{c}{\textbf{AP}} \\
        \cmidrule{4-5} \cmidrule{7-8} \cmidrule{10-11} \cmidrule{13-14} 
        &&& \textit{UDT} & \textit{MDT} && \textit{UDT} & \textit{MDT} && \textit{UDT} & \textit{MDT} && \textit{UDT} & \textit{MDT} \\
        \midrule

acute\_inflammation & 120 & 8 & 1.0 & 1.0 && 1.0 & 1.0 && 1.0 & 1.0 && 1.0 & 1.0 \\
adult & 36631 & 7 & 0.840 & 0.361 && 0.830 & 0.381 && 0.729 & 0.448 && 0.497 & 0.222 \\
arcene & 100 & 10000 & 0.7 & 0.55 && 0.700 & 0.390 && 0.707 & 0.5 && 0.594 & 0.45 \\
arhythmia & 68 & 279 & 0.714 & 0.571 && 0.726 & 0.589 && 0.725 & 0.625 && 0.826 & 0.773 \\
australian\_credit & 690 & 7 & 0.724 & 0.608 && 0.724 & 0.609 && 0.720 & 0.611 && 0.611 & 0.510 \\
balance\_scale & 625 & 4 & 0.872 & 0.856 && 0.857 & 0.866 && 0.473 & 0.602 && 0.08 & 0.120 \\
bank & 45211 & 9 & 0.885 & 0.493 && 0.851 & 0.579 && 0.552 & 0.492 && 0.168 & 0.115 \\
blood & 748 & 3 & 0.773 & 0.493 && 0.727 & 0.524 && 0.575 & 0.400 && 0.306 & 0.218 \\
breast & 683 & 9 & 0.948 & 0.919 && 0.949 & 0.919 && 0.946 & 0.909 && 0.882 & 0.825 \\
car & 1728 & 6 & 0.979 & 0.861 && 0.979 & 0.857 && 0.980 & 0.813 && 0.940 & 0.658 \\
contraceptive & 1473 & 8 & 0.701 & 0.559 && 0.697 & 0.554 && 0.686 & 0.579 && 0.686 & 0.618 \\
compas & 4534 & 12 & 0.985 & 0.886 && 0.985 & 0.863 && 0.975 & 0.671 && 0.923 & 0.438 \\
covertype & 581012 & 54 & 0.808 & 0.507 && 0.809 & 0.502 && 0.796 & 0.460 && 0.640 & 0.348 \\
dexter & 2599 & 20000 & 0.559 & 0.498 && 0.539 & 0.335 && 0.559 & 0.498 && 0.535 & 0.499 \\
electricity & 45312 & 8 & 0.813 & 0.652 && 0.813 & 0.654 && 0.806 & 0.650 && 0.705 & 0.525 \\
fertility & 99 & 6 & 0.85 & 0.8 && 0.875 & 0.837 && 0.916 & 0.888 && 0.4 & 0.333 \\
german & 1000 & 17 & 0.695 & 0.695 && 0.697 & 0.694 && 0.644 & 0.634 && 0.399 & 0.392 \\
gisette & 7000 & 5000 & 0.942 & 0.965 && 0.942 & 0.964 && 0.942 & 0.965 && 0.925 & 0.951 \\
glass & 214 & 9 & 0.860 & 0.930 && 0.860 & 0.923 && 0.462 & 0.654 && 0.069 & 0.213 \\
heart\_failure & 299 & 12 & 0.733 & 0.633 && 0.729 & 0.633 && 0.677 & 0.576 && 0.459 & 0.360 \\
heloc & 10459 & 23 & 0.698 & 0.509 && 0.697 & 0.503 && 0.695 & 0.504 && 0.652 & 0.524 \\
higgs & 98049 & 28 & 0.681 & 0.523 && 0.681 & 0.522 && 0.679 & 0.526 && 0.646 & 0.542 \\
hill & 606 & 100 & 0.5 & 0.606 && 0.394 & 0.606 && 0.500 & 0.606 && 0.5 & 0.563 \\
hypo & 3269 & 23 & 0.909 & 0.640 && 0.897 & 0.719 && 0.610 & 0.576 && 0.177 & 0.102 \\
ipums & 299285 & 8 & 0.949 & 0.640 && 0.937 & 0.733 && 0.633 & 0.560 && 0.248 & 0.071 \\
lrs & 531 & 100 & 0.925 & 0.915 && 0.923 & 0.912 && 0.844 & 0.816 && 0.633 & 0.589 \\
magic & 19020 & 10 & 0.854 & 0.813 && 0.850 & 0.804 && 0.818 & 0.761 && 0.839 & 0.796 \\
madelon & 2000 & 500 & 0.73 & 0.527 && 0.729 & 0.527 && 0.73 & 0.527 && 0.666 & 0.514 \\
house16 & 22784 & 16 & 0.854 & 0.770 && 0.853 & 0.767 && 0.818 & 0.711 && 0.871 & 0.807 \\
ionosphere & 351 & 34 & 0.887 & 0.859 && 0.888 & 0.855 && 0.885 & 0.827 && 0.900 & 0.846 \\
musk & 476 & 166 & 0.802 & 0.854 && 0.800 & 0.854 && 0.792 & 0.851 && 0.704 & 0.767 \\
nbfi & 8308 & 29 & 0.913 & 0.777 && 0.895 & 0.820 && 0.530 & 0.528 && 0.082 & 0.075 \\
ozone & 1847 & 72 & 0.921 & 0.910 && 0.920 & 0.910 && 0.691 & 0.649 && 0.226 & 0.170 \\
page\_blocks & 5473 & 9 & 0.967 & 0.931 && 0.966 & 0.926 && 0.906 & 0.748 && 0.719 & 0.429 \\
phoneme & 5404 & 5 & 0.851 & 0.392 && 0.852 & 0.401 && 0.833 & 0.453 && 0.635 & 0.275 \\
pima & 768 & 8 & 0.714 & 0.370 && 0.702 & 0.368 && 0.656 & 0.408 && 0.477 & 0.316 \\
pol & 15000 & 48 & 0.969 & 0.393 && 0.969 & 0.396 && 0.968 & 0.326 && 0.973 & 0.600 \\
pums & 299285 & 10 & 0.949 & 0.693 && 0.937 & 0.772 && 0.633 & 0.643 && 0.248 & 0.092 \\
planning & 182 & 12 & 0.648 & 0.594 && 0.612 & 0.552 && 0.513 & 0.449 && 0.303 & 0.285 \\
post\_operative & 87 & 8 & 0.722 & 0.5 && 0.677 & 0.525 && 0.561 & 0.469 && 0.322 & 0.266 \\
seeds\_0 & 210 & 7 & 0.833 & 0.976 && 0.822 & 0.975 && 0.767 & 0.964 && 0.650 & 0.952 \\
seeds\_1 & 210 & 7 & 0.976 & 0.976 && 0.976 & 0.975 && 0.982 & 0.964 && 0.933 & 0.952 \\
seeds\_2 & 210 & 7 & 0.928 & 0.952 && 0.927 & 0.952 && 0.910 & 0.946 && 0.838 & 0.886 \\
segment & 2310 & 18 & 0.997 & 0.991 && 0.997 & 0.991 && 0.998 & 0.969 && 0.985 & 0.948 \\
shuttle & 43500 & 9 & 0.999 & 0.999 && 0.999 & 0.999 && 0.998 & 0.998 && 0.999 & 0.999 \\
sonar & 208 & 60 & 0.761 & 0.738 && 0.758 & 0.732 && 0.756 & 0.731 && 0.694 & 0.670 \\
spambase & 4601 & 57 & 0.904 & 0.669 && 0.903 & 0.609 && 0.894 & 0.592 && 0.826 & 0.481 \\
speeddating & 1048 & 62 & 1.0 & 1.0 && 1.0 & 1.0 && 1.0 & 1.0 && 1.0 & 1.0 \\
steel\_plates & 1941 & 27 & 0.915 & 0.915 && 0.916 & 0.917 && 0.740 & 0.754 && 0.296 & 0.309 \\
student\_performance & 1000 & 6 & 0.83 & 0.775 && 0.827 & 0.780 && 0.785 & 0.764 && 0.845 & 0.835 \\
sydt & 9999889 & 8 & 0.979 & 0.672 && 0.978 & 0.704 && 0.956 & 0.705 && 0.979 & 0.875 \\
toxicity & 171 & 1203 & 0.542 & 0.457 && 0.529 & 0.449 && 0.445 & 0.357 && 0.297 & 0.294 \\
twonorm & 7400 & 20 & 0.826 & 0.875 && 0.826 & 0.874 && 0.826 & 0.875 && 0.765 & 0.858 \\
vertebral\_column & 310 & 6 & 0.870 & 0.838 && 0.869 & 0.838 && 0.839 & 0.815 && 0.871 & 0.856 \\
wall\_following & 5456 & 24 & 0.997 & 0.993 && 0.997 & 0.993 && 0.995 & 0.993 && 0.982 & 0.959 \\
wine\_origin & 179 & 13 & 0.916 & 0.944 && 0.917 & 0.944 && 0.916 & 0.937 && 0.803 & 0.868 \\
wine & 6497 & 12 & 0.989 & 0.308 && 0.989 & 0.219 && 0.986 & 0.539 && 0.991 & 0.772 \\
     
        \bottomrule
    \end{tabular}
    \caption{Performance of UDTs and MDTs per dataset. Train-test stratified split $90\%$-$10\%$. }
    \label{tbl:apdx:datasets}
\end{table*}
  
\end{document}